\documentclass[dvipsnames,format=sigconf,anonymous=false,review=false]{acmart}

\AtBeginDocument{%
  \providecommand\BibTeX{{%
    \normalfont B\kern-0.5em{\scshape i\kern-0.25em b}\kern-0.8em\TeX}}}

\copyrightyear{2024} 
\acmYear{2024} 
\setcopyright{rightsretained} 
\acmConference[GECCO '24]{Genetic and Evolutionary Computation Conference}{July 14--18, 2024}{Melbourne, VIC, Australia}
\acmBooktitle{Genetic and Evolutionary Computation Conference (GECCO '24), July 14--18, 2024, Melbourne, VIC, Australia}
\acmDOI{10.1145/3638529.3654138}
\acmISBN{979-8-4007-0494-9/24/07}

\usepackage{lipsum}

\usepackage[export]{adjustbox}[2011/08/13] 
\newcommand\figscale{1.0}

\usepackage{comment}

\begin{document}

\title{Generative Design through Quality-Diversity Data Synthesis and Language Models}

\author{Adam Gaier}
\affiliation{%
  \institution{Autodesk Research}
  \city{Bonn}
  \country{Germany}}
\email{adam.gaier@autodesk.com}

\author{James Stoddart}
\affiliation{%
  \institution{Autodesk Research}
  \city{Atlanta}
  \country{USA}}
\email{james.stoddart@autodesk.com}

\author{Lorenzo Villaggi}
\affiliation{%
  \institution{Autodesk Research}
  \city{New York}
  \country{USA}}
\email{lorenzo.villaggi@autodesk.com}

\author{Shyam Sudhakaran}
\affiliation{%
  \institution{Autodesk Research}
  \city{San Francisco}
  \country{USA}}
\email{shyam.sudhakaran@autodesk.com}


\begin{abstract}
Two fundamental challenges face generative models in engineering applications: the acquisition of high-performing, diverse datasets, and the adherence to precise constraints in generated designs. 
We propose a novel approach combining optimization, constraint satisfaction, and language models to tackle these challenges in architectural design. Our method uses Quality-Diversity (QD) to generate a diverse, high-performing dataset. We then fine-tune a language model with this dataset to generate high-level designs. These designs are then refined into detailed, constraint-compliant layouts using the Wave Function Collapse algorithm.
Our system demonstrates reliable adherence to textual guidance, enabling the generation of layouts with targeted architectural and performance features. Crucially, our results indicate that data synthesized through the evolutionary search of QD not only improves overall model performance but is essential for the model's ability to closely adhere to textual guidance. This improvement underscores the pivotal role evolutionary computation can play in creating the datasets key to training generative models for design. Web article at \textcolor{blue}{\url{https://tilegpt.github.io}}

\end{abstract}


\maketitle
\begin{figure*}
  \includegraphics[width=\figscale\textwidth,center]{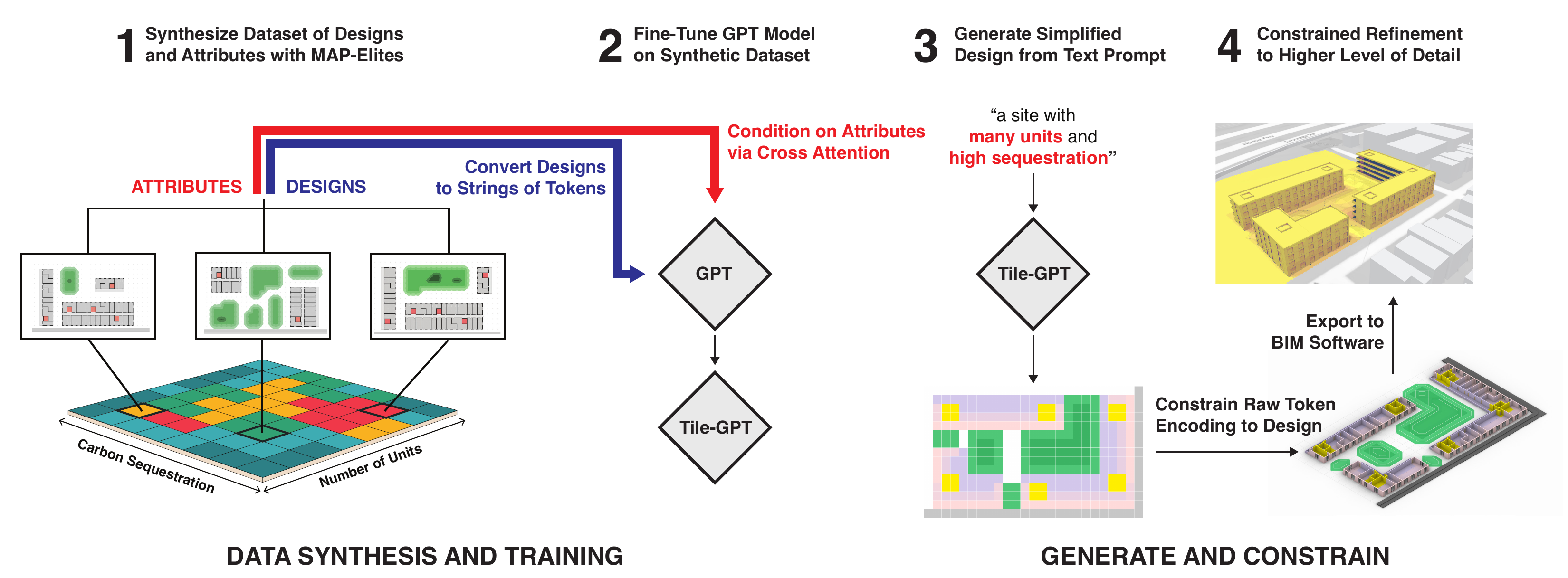}
  \caption
  { 
    Algorithm flow of the proposed generative design approach, TileGPT. (1) A dataset of paired designs and attributes is generated with the MAP-Elites algorithm, which is used to (2) fine-tune a GPT model to produce designs with given attributes. (3) Given a natural language description a simplified design with the described attributes is generated by the GPT model, and (4) given to a constraint satisfaction algorithm, which refines it into a detailed site plan.
  }
  \label{fig:teaser}
\end{figure*}

\section{Introduction}
Generative Design (GD) in architecture represents a paradigm shift in the way designs are conceptualized and realized. It draws inspiration from natural evolution to explore vast design spaces to discover high-performing, innovative solutions \cite{nagy2017beyond}. At its core, GD involves a geometry generator that delineates a broad solution space, coupled with simulations and analytical methods for evaluating each design against a set of metrics. Metaheuristic search algorithms, such as genetic algorithms, navigate this space to identify optimal solutions \cite{nagy2017project}. This approach is versatile and scale-agnostic, making it applicable to a wide range of design problems and scales.

In Architecture, Engineering, and Construction (AEC), GD is most commonly used in the early stages of design~\cite{bradner2014parameters}. This is when the potential to influence outcomes is highest, and the cost implications of design changes are minimal \cite{paulson1976designing,macleamy2004curve}. GD has been successfully applied in numerous AEC projects, enabling practitioners to tackle complex challenges, balance conflicting objectives, and make informed decisions based on solid evidence \cite{nagy2017project,nagy2018generative,villaggi2018generative}. 

But the development and deployment of GD methods requires a high level of technical expertise, which limits their scalability and accessibility. Not only that, the results of GD are large sets of complex solutions, requiring designers to spend as much effort on analysis as on creation. Worse still, typical GD workflows offer limited scope for interaction -- making changes often necessitates rerunning the entire optimization process. 

Large language models (LLMs), which have streamlined many tasks, could also be applied to design. LLMs fine-tuned on labeled segments of existing Mario Bros. levels are able to generate levels which reflect descriptive prompts (e.g., "few enemies," "many pipes")~\cite{mariogpt_gecco, mariogpt_neurips}. Many tasks in architectural design, particularly in the conceptual phase, can be modeled at a similar level of abstraction as video game levels. Experimentation has already begun in AEC to take advantage of the same tile-based layouts and procedural content generation (PCG) techniques used in games.~\cite{wfc_qd, wfc_fos}.

Crucially, to adapt an LLM-based approach to design in this way it is necessary to have a large corpus of labeled data.
Quality-Diversity (QD)~\cite{pugh_qd,cully_qd} approaches are able to generate large collection of solutions, ideal for use as training data. These high performing collections of span user-defined features, allowing users to define the design features to explore, and then generate a bespoke dataset that spans those features.

Designs generated by language model are created only through the learned statistical relationships, but in design it is necessary that constraints are followed. Rather than forcing the LLM to learn every constraint, we instead task it with creating a conceptual plan which is then handed to the Wave Function Collapse (WFC) algorithm~\cite{wfc}, a PCG approach based on constraint satisfaction~\cite{wfc_constraint_solving}.

The language model interprets the designers intent through natural language prompts, and generates designs at a high level, such as the placement of buildings and green spaces. These designs are then processed by WFC to generate the detailed layout of modules. This method ensures that the final design not only resonates with the input provided by the designer but also rigorously complies with the constraints of construction.

The outlined system, dubbed TileGPT, demonstrates:
\begin{itemize}
  \item The integration of QD with PCG techniques to synthesize tailored labeled datasets of high performing solutions.
  \item The use of a fine-tuned LLM to interpret and implement design directives in natural language and apply them to a real-world generative design case.
  \item The application of constraint satisfaction to guarantee the validity of designs generated by an LLM.
\end{itemize}

This novel approach integrates QD, LLMs, and constraint satisfaction within the GD framework. This integration aims to enhance the accessibility of GD methods, reduce the technical barriers to their use, and provide more intuitive, interactive design manipulation capabilities through natural language inputs.

\section{Background}
	\subsection{Wave Function Collapse}\label{sec:wfc}	
	    Wave Function Collapse (WFC) is a procedural content generation technique, popularized by Maxim Guman \cite{wfc} for creating 2D and 3D content, in the form of a constraint satisfaction algorithm. It is similar to the Example-Based Model Synthesis \cite{model_synthesis} method and is adept at generating non-tiling, self-similar structured data based on sparse input examples. 

The algorithm works through iterations of a single cell collapse — assignment to a single fixed state — and neighborhood propagation — where surrounding tiles are constrained to compatible patterns with the collapsed cell. Cells are collapsed in order of minimum entropy, measured as the certainty of a specific outcome from the weightings of potential states, precisely defined as:



\begin{equation}
    \text{Shannon Entropy} = \log\left(\sum w_i\right) - \frac{\sum(w_i \times \log(w_i))}{\sum w_i}
\end{equation}
where $w_i$ represents the weight of each potential state for a cell. The weight reflects the likelihood or frequency of a particular state occurring based on the adjacency constraints and the neighbors.

The WFC methodology consists of a four-step process:

\begin{enumerate}
    \item \textit{Pattern extraction}: Utilizing one or more self-similar input examples, WFC identifies cell adjacencies, which are used to form a domain of possible constrained states. 
    \item  \textit{Initialization and pre-constraint}: The output is initialized with each cell represented as an array of potential states. Individual cells can be pre-constrained to a subset of these states. This is commonly used to enforce boundary conditions or enable controlled generation from an initial pattern.
    \item \textit{Cell collapse}: The output cell with the lowest entropy value is selected for collapse. From the selected cell's possible states, a single, final state is chosen using a weighted random selection and all other potential states discarded. In the event more than one cell has the same lowest entropy value, the cell to collapse is chosen randomly from the candidates.
    \item \textit{Propagation}: After a cell is collapsed, the solver iterates through the adjacent cells and removes all pattern states incompatible with collapsed cell.
\end{enumerate}

The solver repeats steps 3 and 4 until the output is fully collapsed, with each cell assigned to a single state, or until a contradiction arises, indicating that the solver cannot satisfy all constraints.

	\subsection{Quality-Diversity}\label{sec:qd}	

Quality-Diversity (QD) approaches, like MAP-Elites~\cite{mapelites,me_nature}, search for high-performing solutions which cover a range of user-defined features. Generating diversity along features rather than only objectives makes QD well suited to the needs of GD, as designers are often interested in other attributes beyond objectives~\cite{bradner2014parameters}. QD has been applied in various design domains including aerodynamics~\cite{gaier2017aerodynamic,produqd,sphen}, game design~\cite{alvarez2019empowering,gravina2019procedural,gonzalez2020finding} and architecture~\cite{galanos2021arch,wfc_qd, tdomino}. 

QD produces a collection of solutions in a single run. The ability to produce numerous high-quality and varied solutions positions QD as an ideal tool for synthesizing datasets for machine learning. Collections of solutions generated through QD have been effectively used in creating surrogate models that predict performance~\cite{sail_ecj, bopelites, dsme}, building generative models to aid optimization~\cite{dde, pms} and exploration~\cite{stepping, taxons, aurora}, creating conditioned reinforcement learning policies~\cite{faldor2023map}, and fine-tuning language models to produce virtual creature body plans~\cite{elm}. In this work, we leverage the designs produced by MAP-Elites to fine-tune and condition a language model to produce designs based on text prompts.
 	\subsection{Language Models}\label{sec:lm}	

Large Language models (LLMs) are powerful and versatile, able to learn from massive datasets for sequence modelling tasks such as generating text~\cite{brown2020language}, code~\cite{GitHubCopilot, li2023starcoder}, and multimodal outputs such as images and robot states~\cite{driess2023palm}. These models leverage attention mechanisms \cite{vaswani2023attention} to capture patterns in long term sequences. Pretrained LLMs can be fine-tuned for diverse downstream sequence modelling tasks, reusing the models parameters as a starting point and adding an additional layer trained from scratch. These tasks are not limited to text, but can be generalized to other sequences, such as tile based layouts. Several works have explored this in the context of video games, including MarioGPT, a fine tuned LLM for Mario level generation \cite{mariogpt_neurips, mariogpt_gecco}. The authors showed that MarioGPT was able to generate coherent and playable levels whose layout could be guided by text -- an approach we build upon in this work.


\section{Method}

The TileGPT system, described in detail below, proceeds as follows:

\begin{enumerate}
    \item Dataset Generation
      \begin{enumerate}
        \item A dataset is generated by using MAP-Elites to search the space of designs that can be generated by WFC.
        \item Each design in the dataset is paired with a text label such as 'many units' or 'little carbon sequestration' based on the features of the design.
      \end{enumerate}
    \item Model Training
      \begin{enumerate}
        \item A GPT model is fine-tuned using this dataset of designs, adapting it to produce layouts one tile at a time.
        \item Attribute labels are converted into numerical vectors via a text encoder and incorporated into the fine-tuning process through a cross attention layer.
      \end{enumerate} 
    \item Layout Generation
      \begin{enumerate}
        \item The GPT model is provided with a natural language prompt corresponding to the desired attributes and generates a design at a low level of detail.
        \item These rough layouts are refined by the WFC algorithm, ensuring local constraint satisfaction and validity.
      \end{enumerate}     
\end{enumerate}

        \subsection{Dataset Generation}\label{sec:datagen}	
            \subsubsection*{Synthesizing Data with MAP-Elites}
\begin{figure}[t]
  \includegraphics[width=\columnwidth,center]{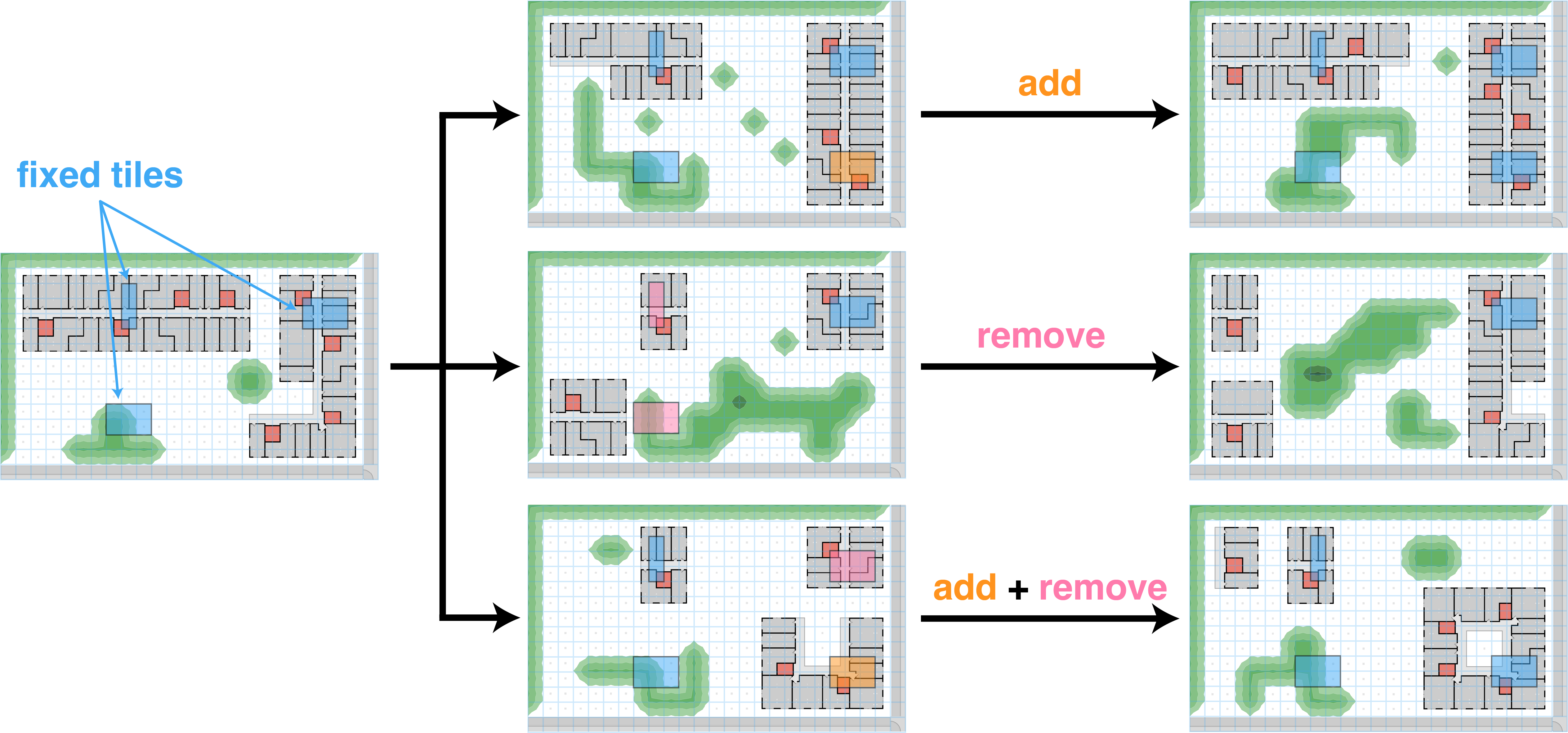}
  \caption
  { 
  Mutation of a WFC genome. Fixed tiles are encoded into the genome, and set at the start of a WFC rollout, influencing the development of the final design.
  }
  \label{fig:evo_wfc}
\end{figure}    
WFC stands out for its ability to generate a wide array of unique designs from minimal initial examples, a potential we leverage for the generation of synthetic datasets. By conducting numerous iterations or `rollouts' of the WFC algorithm, a large volume of data can easily synthesized. 

Despite its versatility, WFC-generated designs are not ideal samples, particularly when performance of the designs is a priority. In a domain like site design, common issues with WFC include inefficient utilization of space, which could be better employed for buildings or landscaping. Furthermore, there's a tendency for the attributes of the designs to converge towards average values, leading to a dataset that lacks extremes and, as a result, limits the scope of what models can generate. The sparsity of varied and compelling examples in the dataset restricts the model's ability to produce innovative designs or to respond appropriately to text prompts.

To overcome challenges related to the quality and uniformity in design generation, we use MAP-Elites for data synthesis. By adopting a diversity-based optimization strategy, we actively seek out high-quality solutions that encompass a broad spectrum of features. This method moves beyond simply sampling, ensuring the creation of a dataset that is both diverse and of high quality. The enriched dataset thus obtained is pivotal in training our model, enabling it to produce designs that are not only varied but also superior in quality. This refined approach significantly boosts the model's capability to generate diverse site layouts, enhancing the overall effectiveness of the design process.

\subsubsection*{Optimization of Designs with WFC}
To use WFC in the optimization process, we must devise a way of effectively searching the space of WFC produced designs.
The core strength of WFC is that it is capable of producing a large variety of solutions that follow a consistent style and set of constraints.
This constrained expressivity makes it an appealing option for optimization, but searching the space of solutions through WFC is challenging. The variety in WFC comes from the chaotic elements of its generation process -- small changes in the initial conditions or early choices have dramatic consequences for the final result.

The a common lever  to guide WFC is to adjust the probability of each tile being chosen when a cell is `collapsed'. Macro level differences are possible to induce in this way, but it is impossible to replicate or preserve distinct tile patterns. 
Adjusting tile weights alone does not produce a suitable encoding for optimization. An encoding based on a tile weight genotype and fully collapsed tile phenotype is highly non-local~\cite{locality} -- a small change in the genotype produces a large and unexpected change in the phenotype, dooming any search algorithm to be little better than random. 

We can consider the mapping of genotype to phenotype through the intermediary of WFC as a \textit{probabilistic} encoding, where each genome maps to a distribution of phenotypes. To create an encoding which is more local, and so more amenable to search, we must narrow this distribution while also making it heritable. 

At the start of the WFC algorithm we can fix a set of tiles, preserving a few existing parts of the parent design and allow the algorithm to generate the remainder. These fixed tiles can be included as part of the genome and passed on to child solutions. A genome composed of fixed tiles and tile weights is a more local encoding -- children resemble parents, and small changes in genotype typically produce small changes in phenotype. The more tiles which are fixed the narrow the distribution of possible mappings from genotype to phenotype.

We can further instantiate individuals by including a random seed, ensuring that a given genome always produces the same phenotype. The resulting genotype is represented as a tuple comprising tile weights, fixed tiles, and a seed. It takes the form:

\[
\text{Genotype} = (T_{\text{weight}}, T_{\text{fixed}}, \text{Seed})
\]

Where, \( T_{\text{weight}} \) is a vector of tile weights, \( T_{\text{fixed}} \) is a list of tuples with each tuple representing a tile type and its position in the grid.

To search this space, we apply a mutation operator, which involves the following steps:

\begin{enumerate}
    \item The  \( T_{\text{weight}} \)  vector is modified by the addition of Gaussian noise, adjusting the weights either upward or downward.
    \item Tiles are added or removed from the \( T_{\text{fixed}} \) list.
    \item $\text{Seed}$ is reset to a new random integer.
\end{enumerate}

At each generation an equal number of individuals are chosen to have tiles removed and added. Tiles are chosen to be added or removed randomly, and the number added or removed drawn from a uniform distribution between 1 and 4 tiles. 

Fixed tiles are added from the phenotype of the parent solution. Adding tiles in this way not only allows children to inherit the  same structures, it ensures that the constellation of fixed tiles is a valid one -- we know there must be at least one valid phenotype to be found by WFC with that set of tiles. The process of fixed tile mutation is illustrated in Figure \ref{fig:evo_wfc}.

The iterative adding and removing of tiles allows a search algorithm to purposefully search through the space of designs generated by WFC -- designs which are guaranteed to follow the guidelines and requirements of the designer.

\begin{figure}[t]
  \includegraphics[width=\columnwidth,center]{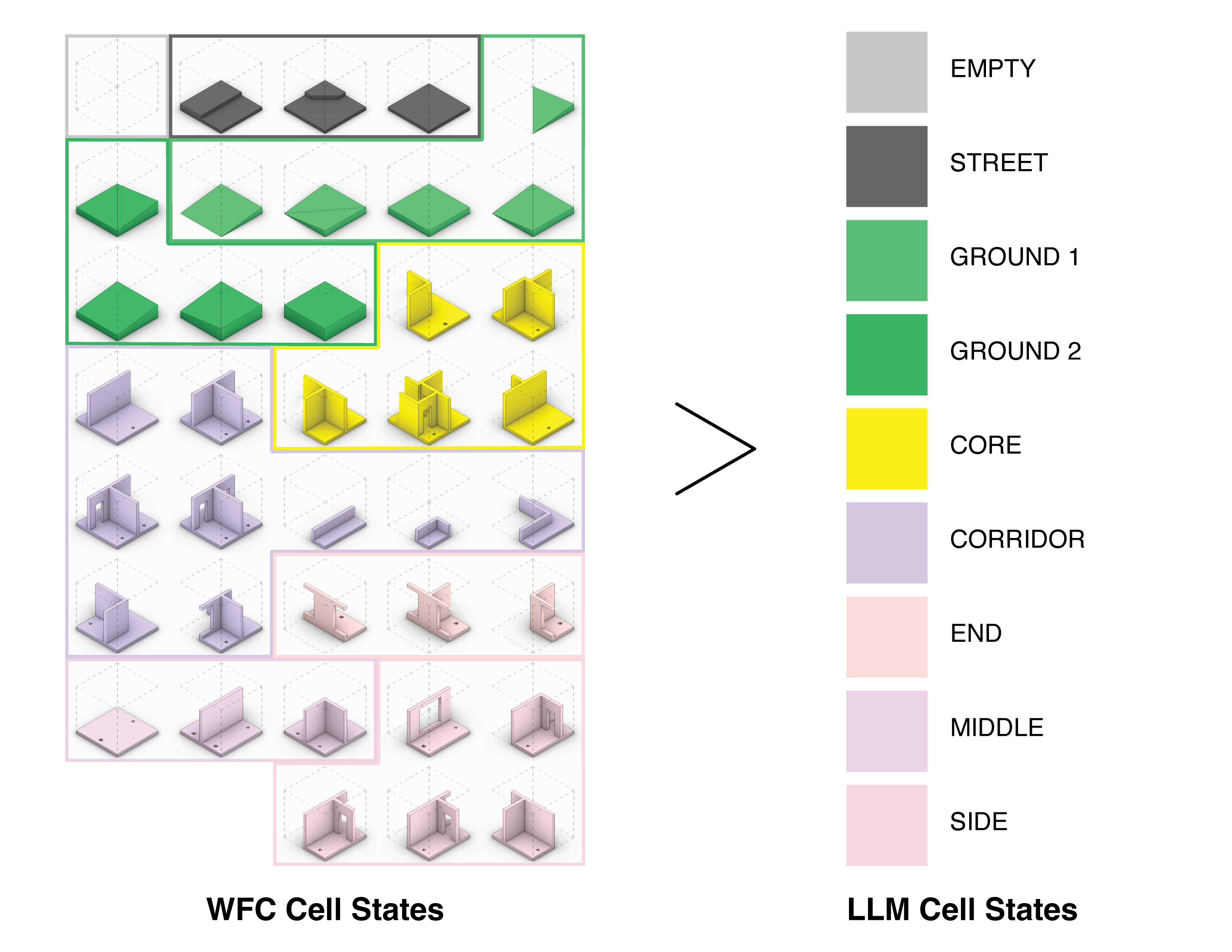}
  \caption
  { 
    Possible WFC cell states and their simplifications for tokenization. Designs are evaluated using the WFC cell states, but generated using the reduced set of LLM cell states.
  }
  \label{fig:tileset}
\end{figure}
\begin{figure*}[t!]
  \includegraphics[width=\figscale\textwidth,center]{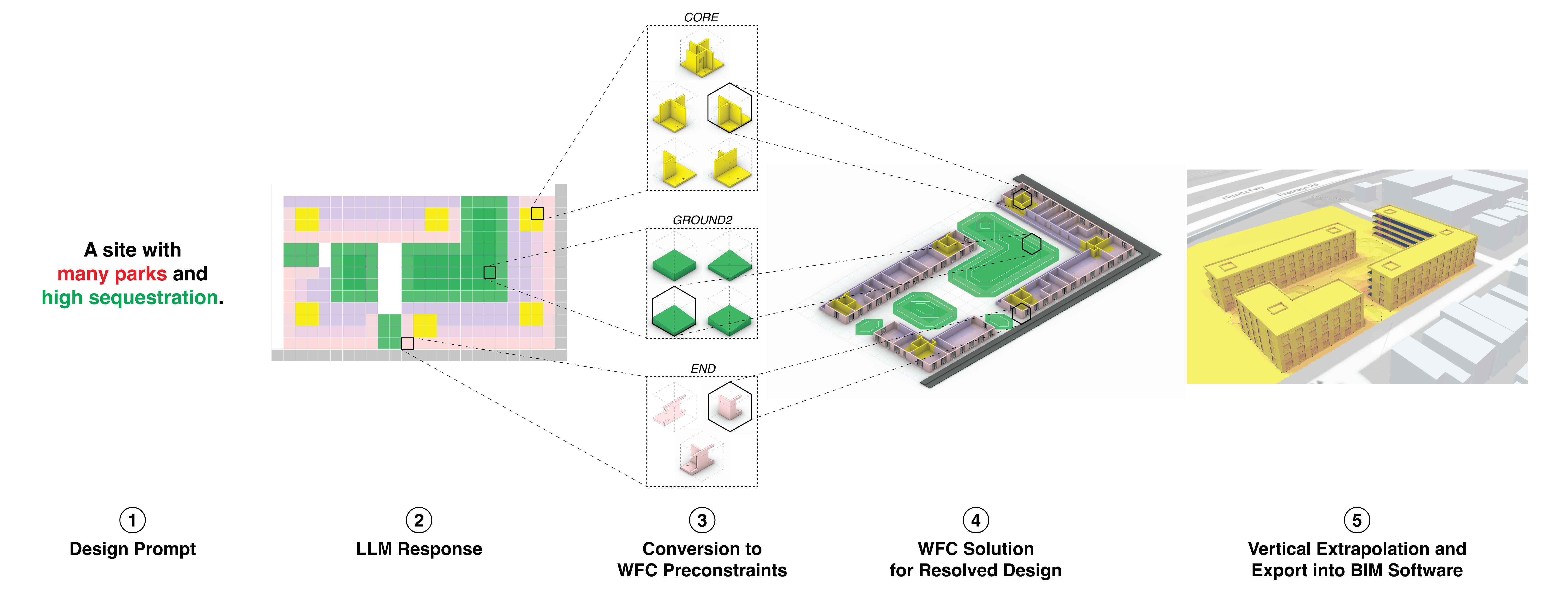}
  \caption
  { 
    Layout Generation in TileGPT. (1) A site description is provided to the model, which (2) produces a high level layout. (3) This layout is converted into preconstraints for the WFC algorithm, which (4) generates detailed geometry. The 2D geometry can be then be extruded (5) into a form suitable for use with commercial design software.
  }
  \label{fig:generate}
\end{figure*}

\subsubsection*{Dataset Preprocessing}
The number of potential tiles, considering their rotations and reflections, can easily reach into the hundreds -- and each tile comes with its own unique set of adjacency rules\footnote{see Appendix \ref{sec:tile_encoding} for the full set of 216 tiles in our application}. Training a GPT model to predict tokens at this level of granularity distracts from its central objective: facilitating global-level optimization and exploration.

Our approach positions the GPT model as a strategic director in the design process. Its role is not to micromanage the minutiae of tile adjacencies but to guide overarching design decisions. This perspective aligns the model’s strengths with the demands of high-level conceptual design, and steers clear of the intricacies of individual tile relationships.
To streamline this process, we categorize the full tile set into a smaller set of distinct functional groups, as illustrated in Figure \ref{fig:tileset}. This categorization substantially reduces the complexity the GPT model has to manage.

Designs are represented as a grid of tiles, but to convert these designs into tokens we transform each into one of these functional categories. Subsequently, each category is represented by a unique character (e.g., 'A', 'B', 'C'). We then flatten this grid of characters into a vector format to fit the standard sequence completion training paradigm of GPT models.
Each site’s features -- defined by their coordinates in the MAP-Elites grid -- are paired with their respective design. These are then translated into high-level natural language descriptions during training (e.g., "few/some/many parks").

	\subsection{Language Model Training}\label{sec:training}	

\begin{figure}[ht]
  \includegraphics[width=\columnwidth,center]{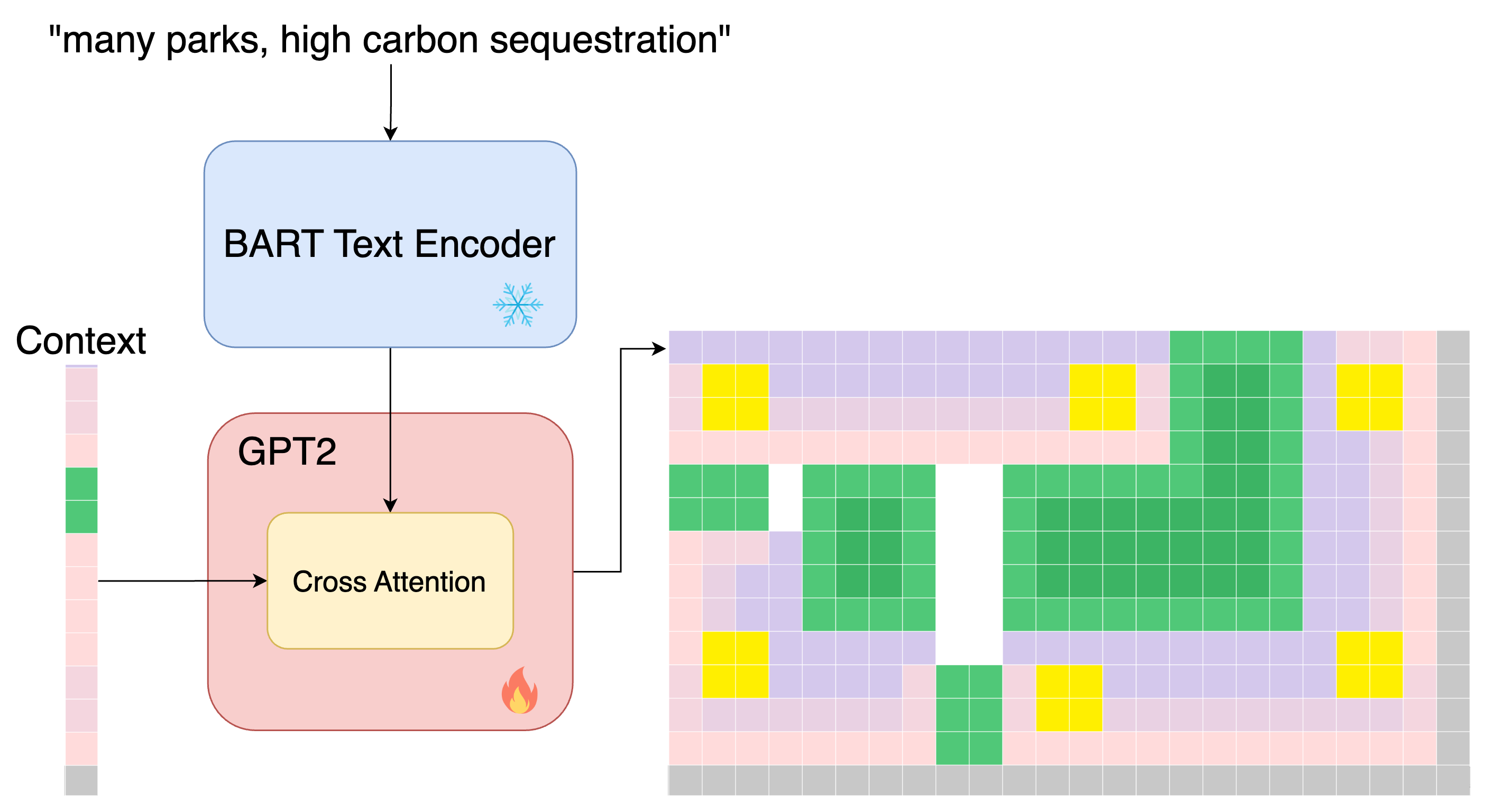}
  \caption
  { 
    TileGPT architecture. Text prompts are encoded through a frozen text encoder and are combined with previous tiles in GPT2's cross attention mechanism.
  }
  \label{fig:model}
\end{figure}

A causal language model is finetuned to learns "next tile prediction", analogous to the "next token prediction" objective which most causal language models are optimized for -- the model learns to generate a design by predicting a single tile based on a sequence of previous tiles. Previous work has shown that by finetuning LLMs for tile generation they can generate new playable levels in Sokoban\cite{sokoban_gpt} and Mario Bros\cite{mariogpt_neurips}. Similar to \cite{mariogpt_neurips}, we choose a distilled version of GPT2 (DistilGPT2) \cite{sanh2020distilbert} as our base LLM to finetune, with additional cross attention weights used for prompt conditioning. To incorporate these prompts, we utilize a frozen text encoder (BART) \cite{lewis2019bart} to embed the prompts as a vector of floats. These vectors are averaged and used in the cross attention weights in combination with the encoded tile sequence. All previous tiles are used as context for predicting the next tile. The architecture is illustrated in Figure \ref{fig:model}.

Because we use DistilGPT2, the model in TileGPT is relatively small and utilizes only 96 million trainable parameters. This allows for training efficiently on a single GPU. We train TileGPT for 500,000 steps, sampling 16 random designs uniformly at each training iteration and optimize weights using Adam optimizer \cite{kingma2017adam}.

        \begin{figure*}[t]
  \includegraphics[width=\figscale\textwidth,center]{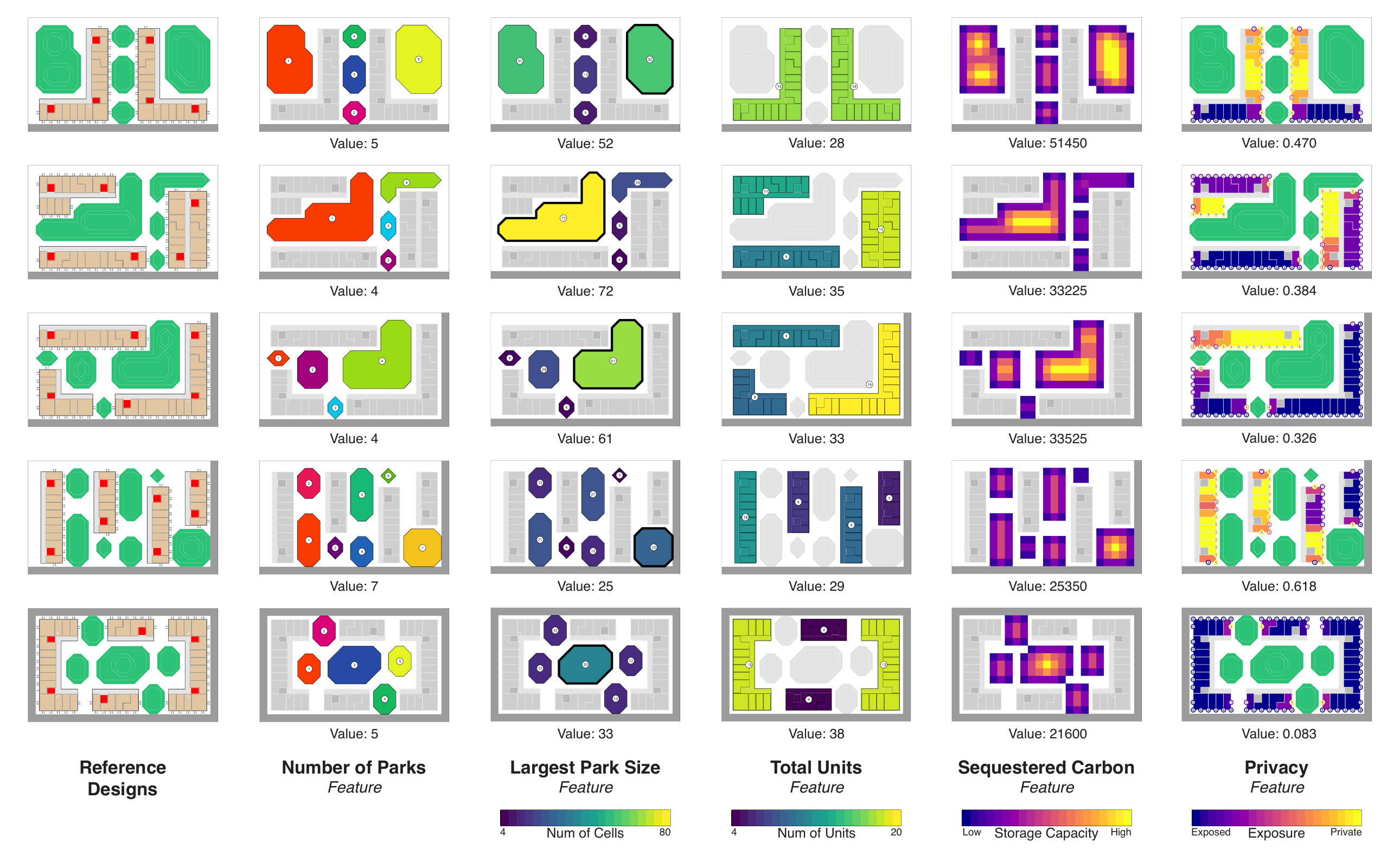}
  \caption
  { 
  Features explored with MAP-Elites. Layouts which span these features are generated to form a dataset for training. 
  }
  \label{fig:metrics}
\end{figure*}	  
	\subsection{Layout Generation}\label{sec:generation}	
            
To use the model for design generation, we follow the a series of steps, as depicted in Figure \ref{fig:generate}. In this integrated process, the GPT model lays the foundation for the overarching design based on natural language prompts, while WFC ensures its practical feasibility and completeness.

\paragraph{Step 1: Design Initiation via Prompt}
The process begins with the input of a design prompt. This prompt incorporates the natural language parameters our model has been trained on. The system inserts randomly sampled prompts for those not provided. These prompts are converted to a vector and used as a constant input to the cross-attention layer -- laying the groundwork for the subsequent design generation.

\paragraph{Step 2: LLM-Driven Site Design Formation}
Following the initial prompt, the GPT model, steered by the textual input, engages in an iterative process of selecting tiles from a simplified set of categories. These selections form a high-level blueprint, outlining the fundamental structure of the design.

\paragraph{Step 3: Translation to Permissible Tile Sets}
The basic tile types delineated by the GPT model are then transformed into a set of allowable tiles. For instance, a `building core' might be represented in every possible orientation. This step refines the blueprint, preparing it for more detailed procedural generation.

\paragraph{Step 4: Detailed Design Completion through WFC}
The refined blueprint is subsequently transferred to the WFC. WFC selects from a comprehensive tile-set to add intricate details, from orientations to placement of windows and interior walls.

\paragraph{Step 5: Finalizing a Valid Design}
Upon completion of WFC, we obtain a single, valid design. This design is not only complete in its structure but also readily transferable to Building Information Modeling (BIM) software for detailed editing and analysis.

\vspace{\baselineskip}
Importantly, this procedure is not rigid. Users have the flexibility to modify the design iteratively. For example, a portion of the site can be erased and re-generated by inputting an alternative text prompt, directing the system to refill the area with a design that incorporates specific desired features. This iterative capability enhances the adaptability and user-interactivity of our design generation process.

\section{Experiments}
	\subsection{Setup}\label{sec:setup}	
            
We test our system in a real-world design scenario: the design of apartment complex layouts for prefabricated housing. As part of an applied research collaboration with the modular construction company FactoryOS\footnote{\url{https://factoryos.com/}}, we derived our modules from their real-world catalog of prefabricated apartment units and worked together to test the WFC algorithm for early stage design.

Adjacency rules for our WFC algorithm are derived from a small set of manually created reference designs (see Appendix \ref{sec:wfc_base}). Each generated site layout consists of a 25x15 grid, totaling 375 tiles. These tiles represent various elements: livable building component modules, utility elements like corridors and cores, more or less intensive landscaping such as trees or lawn, and unused spaces and streets. Site borders are fixed, surrounded by street or landscaping.

Sites are evaluated on five metrics: number of parks, largest park size, total units, sequestered carbon, and privacy, each illustrated in Figure 6. A site's performance is gauged by the proportion of non-empty tiles. Each site is labeled with a text prompt that mirrors these features, divided into low, medium, and high values, for a total of 243 ($3^5$) possible text labels. For clarity we will refer to these metrics as `features' and an instance of these features as an attribute (high privacy vs. low privacy).

	\begin{figure*}[t]
  \includegraphics[width=\figscale\textwidth,center]{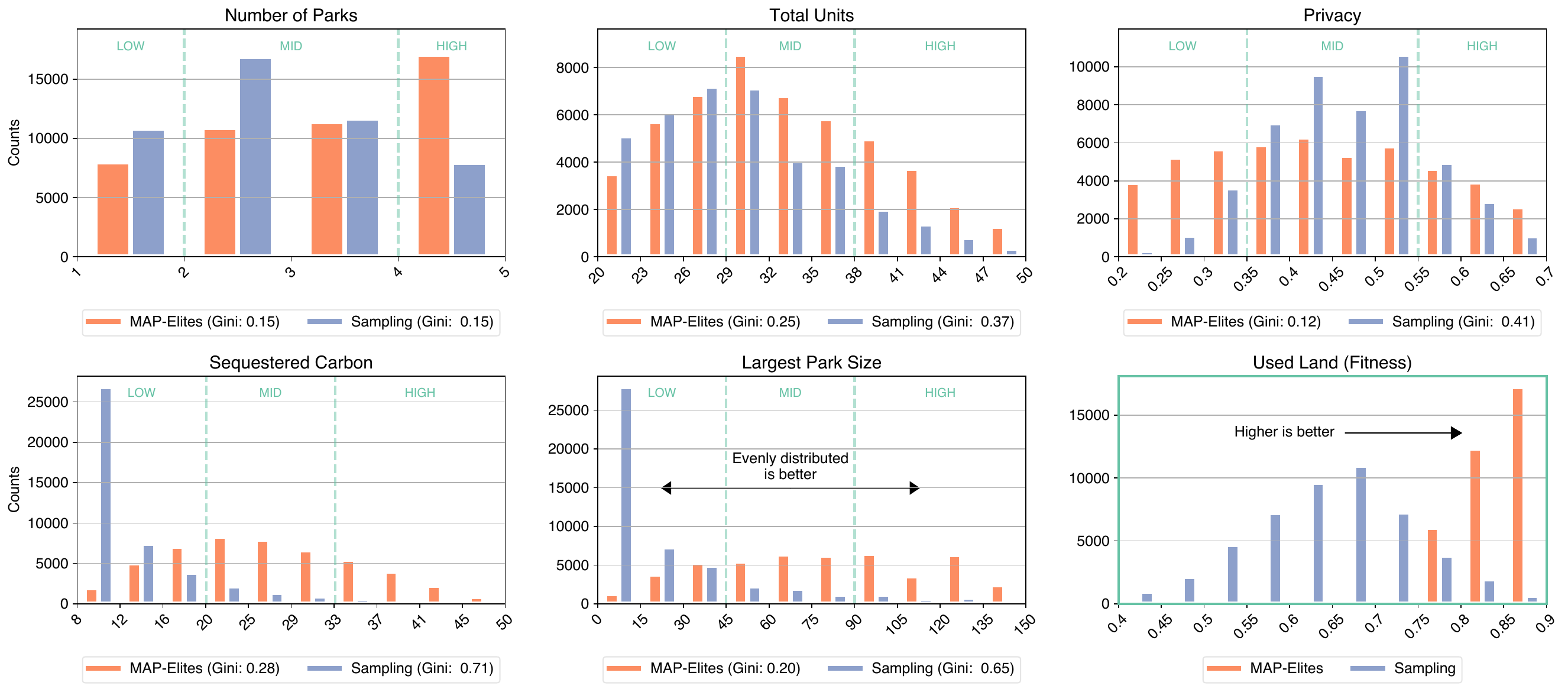}
  \caption
  { 
  Distribution of feature and performance values of in datasets of designs generated with MAP-Elites and Sampling. Gini coefficient of number of samples in each bin is provided to aid interpretation of distributions. Demarcation of the qualitative labels used to train the model (e.g. low, mid, high number of units) show in green.
  }
  \label{fig:data_dist}
\end{figure*}
        \subsection{Results}\label{sec:results}	
        \subsubsection*{Experiment Objectives and Methodology}
Experiments are designed to evaluate our system, a language model fine-tuned on a synthetic dataset, in generating designs that are then refined to meet specific constraints and criteria. We focus on two key aspects:

\begin{enumerate}
    \item \textbf{Validity:} Does our system reliably produce valid designs that can be transformed into complete layouts by WFC?
    \item \textbf{Fidelity:} How well do designs align with the given prompts?
\end{enumerate}

Where a layout is `complete' if it is filled with a set of tiles that obey all adjacency constraints, and a design is considered to `align' with the prompt if the attribute value is in the ranges defined during training for each text prompt (see Figure \ref{fig:data_dist} for demarcations). We evaluate our model with the following exhaustive approach:
\begin{itemize}
    \item We prompt the model to produce 100 designs for every combination of prompts, amounting to 243 prompts. 
    \item Validity is measured by the WFC solver's ability to generate a complete layout from each design.
    \item Fidelity is measured for each valid layout, with achieving fidelity when the attributes match those specified in the prompt, each attribute evaluated separately.
\end{itemize}
We investigate the impact of employing a QD approach in the generation of the synthetic dataset. Two datasets are used to produce models, one generated with MAP-Elites and the other by sampling WFC, each with a dataset contains a total of 50,000 designs each.

\subsubsection*{Comparative Analysis of Datasets}
It is informative to first examine the differences in the datasets generated by sampling and by MAP-Elites. Analyzing the composition of these datasets provides a clearer understanding of the differences in the resulting models.

A key aspect of producing expressive models is ensuring a diverse range of features in the dataset. Ideally, this would manifest as a uniform distribution across all features. While a completely filled MAP-Elites archive would produce this ideal scenario, in practice there are inherent trade-offs in features, and not every combination can be produced, so creating some imbalance in unavoidable.

The distribution of feature values in the designs of each dataset is shown in Figure \ref{fig:data_dist}. To underline the difference in uniformity, we also calculate the Gini coefficient\footnote{A Gini coefficient of 0 indicates perfect uniformity, while 1 indicates all samples concentrated in a single bin}, a measure of inequality, of the number of samples in each bin.

This analysis reveals that MAP-Elites produces a far more uniformly distributed dataset compared with sampling. More than half of all samples generated by sampling WFC are in the lower tenth of sequestered carbon and large parks -- randomly generated designs rarely yield large parks, which are crucial for substantial carbon sequestration. Random sampling simply cannot reliably cover the extremes of some feature distributions.

In addition we examine the distribution of performance values (Figure \ref{fig:data_dist}, bottom right). The datasets generated by sampling alone tend to follow a normal distribution around a low mean. In contrast, MAP-Elites actively seeks out high-performing designs. This distinction underscores the effectiveness of targeted search methods like MAP-Elites in creating datasets that not only span a broad feature range but also include high-performance design options, which are less likely to emerge through random generation.

\begin{figure}[htbp]
  \includegraphics[width=1.0\columnwidth]{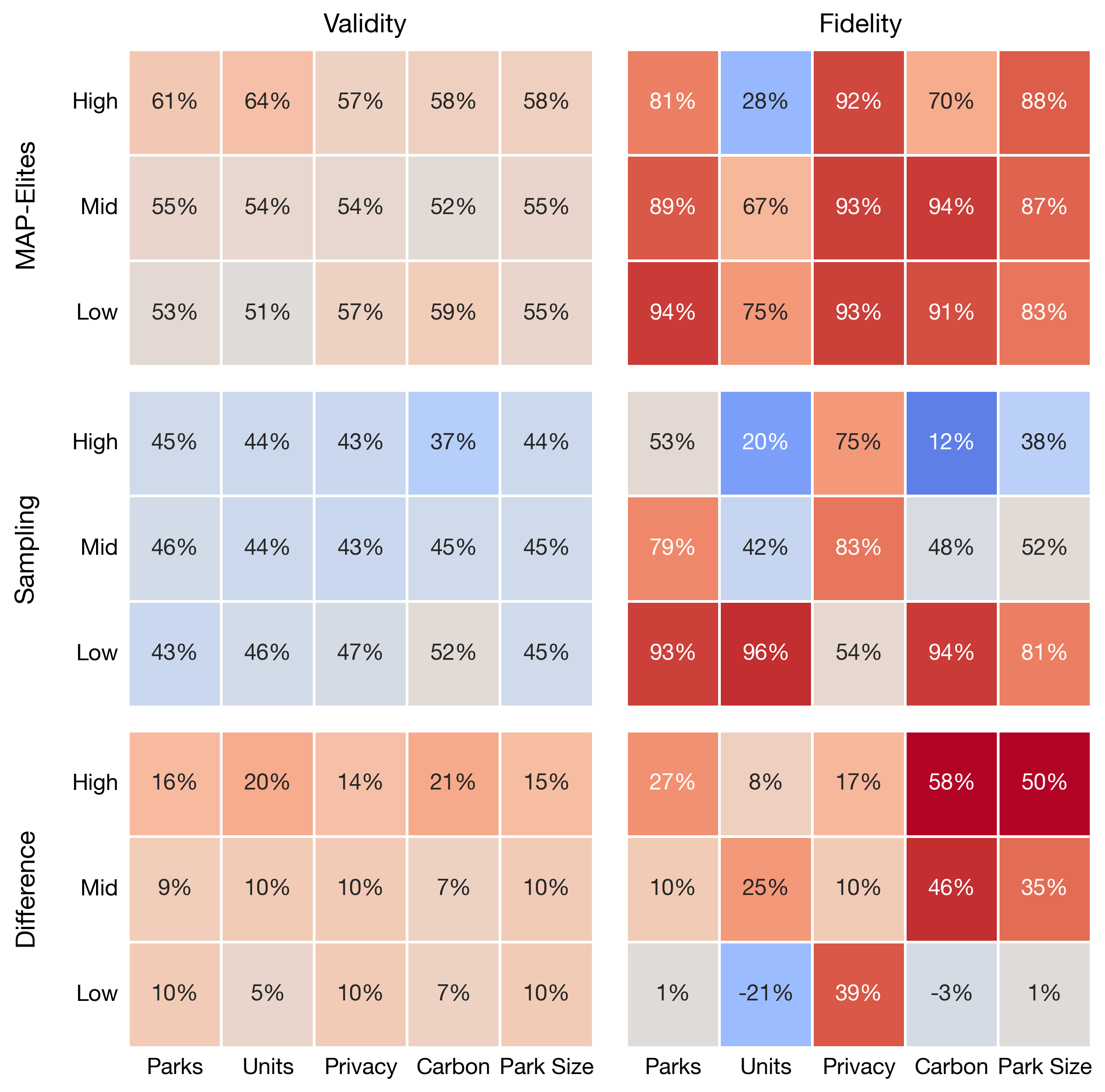}
  \caption
  { 
  Model performance when trained on a MAP-Elites synthesized dataset vs. one obtained by sampling. Each cell represents the mean of a single prompt (e.g. "High number of parks") in combination with every other prompt (varied levels of units, privacy, carbon, park size). \textit{Validity}: how often a design with this prompted feature generates a valid design. \textit{Fidelity}: how many valid solutions follow the prompt.
  }
  \label{fig:model_perf}
\end{figure}

\paragraph{Model Performance}
The performance of each model, including the differences between them, is shown in Figure \ref{fig:model_perf}. The model trained with MAP-Elites synthesized dataset demonstrates a higher level of fidelity to the design prompts across nearly all categories. Though the category of 'total units' shows comparatively weaker performance, this can be attributed to the model's limited control over this aspect; while it can outline the building design, the actual generation of walls—and consequently the number of units—is determined by the WFC solver and randomness of the seed.

The model trained on the dataset generated by WFC exhibits uneven performance, mirroring the inconsistencies in its training dataset. The model struggles to generate designs with high carbon sequestration, large parks, or low privacy solutions, all of which are underrepresented in the sampled dataset. For attributes with abundant data, such as low carbon sequestration or number of parks, the model performs well. That the sampled dataset is lower-performing, with a lot of empty tiles, translates into fewer and smaller parks, and fewer units. This alone may be enough to bias the generation toward these attributes, regardless of the prompt.

The validity of designs generated by the WFC-trained model is lower across all categories, particularly in the 'high' level categories where the fidelity is also lacking. This trend can be attributed to the model's limited exposure to the cross-attention signal of rarer prompts in the WFC dataset, leading to challenges in handling less predictable inputs and consequently producing invalid designs.

The results underscore that the caliber and variety of the training data are key to successful model training. In particular this emphasizes the superiority of QD methods in creating rich and varied datasets, proving their effectiveness for sophisticated, real-world design problems where random sampling is not sufficient.

\section{Discussion}
This work introduces a novel approach to generative design, addressing the challenges of data availability, ease of use, and constraint compliance. Our method combines optimization techniques, constraint satisfaction mechanisms, and the generative capabilities of language models to remedy stubborn difficulties intrinsic to GD.

Building on existing generative design methods, our approach transforms their main weakness—the overwhelming volume of results—into a key advantage. Instead of requiring users to sift through thousands of generative design outcomes, these results become raw material to train a model to help them explore the possibilities of design. This integration allows users direct access to the exploratory benefits of evolutionary AI and the precision of constraint-satisfying symbolic AI, all through the user-friendly interface of a generative AI language model.

Our current system was built on simple tile representations, and while many layout problems in architecture can be encoded in this way, it is an obvious limitation to the technique's versatility. Alternative tokenization schemes would enable the generation of different geometries, and many such approaches are already gaining traction for manufacturing design~\cite{xu2022skexgen,xu2023hierarchical,jayaraman2022solidgen}

The conditioning of the model on features is currently based on linear ranges of user-defined features; however, future implementations could utilize non-linear regions or integrate more descriptive natural language labels for more intuitive exploration. Approaches like Quality-Diversity with AI Feedback~\cite{qdaif}, especially combined with multimodal models which could automatically label site plans with more qualitative attributes, could further enhance the system’s capability for generating intuitive and meaningful design features.

Although not explicitly evaluated in this paper, the system is designed to be interactive. Users can modify specific areas of a site layout according to their prompts, enabling high-level exploration and alteration of site plans
Such prompt-guided changes can act as high-level mutation operators, as shown in MarioGPT~\cite{mariogpt_neurips}, offering a novel avenue for interactive and dynamic design modification.

Beyond the specifics of the presented system, this work represents a broader approach for applying generative models in engineering and architecture. This approach rests on three pillars: diversity-based optimization for generating high-quality datasets, the use of large models for generation and interaction, and constraint satisfaction algorithms that take the final step in generation to ensuring the valid designs. By weaving a generative model into the fabric of the design process, we mitigate the need for extensive post-hoc analysis typically associated with generative design. Instead, we pave a path for purposeful exploration, allowing for both controlled directives and serendipitous design outcomes. 

\bibliographystyle{template/ACM-Reference-Format}
\bibliography{bib/qd,bib/ml,bib/gd}
\newpage
\onecolumn
\appendix
\section*{Appendix}

\section{Source Code}
Source code can be found linked from the project's permanent home: 
\textcolor{blue}{\url{https://tilegpt.github.io}}

\section{WFC Base Designs}

\label{sec:wfc_base}
    \begin{table*}[h!]
    \centering
    \begin{tabular}{c c c c c}
        \includegraphics[width=0.18\linewidth]{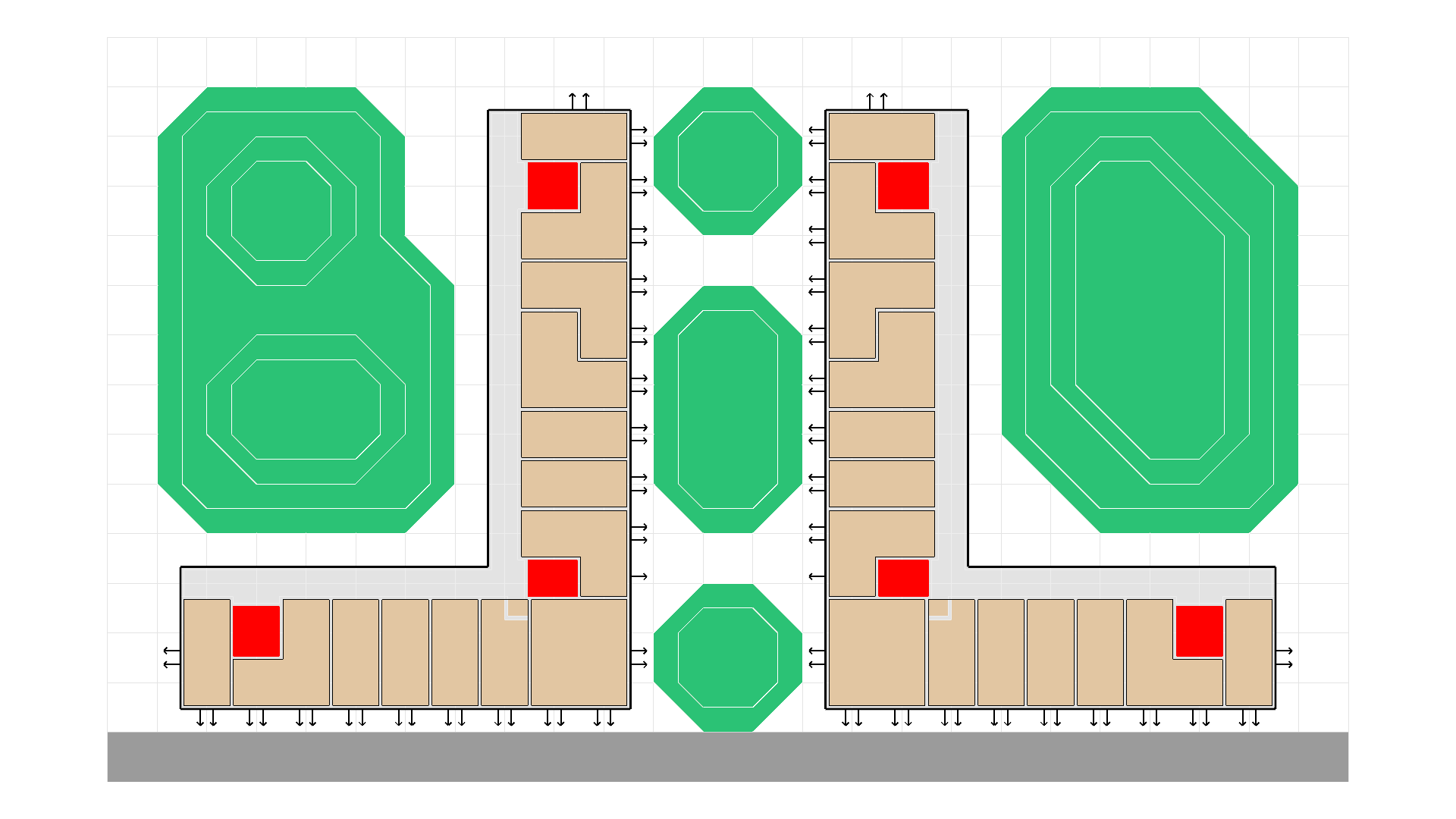} & 
        \includegraphics[width=0.18\linewidth]{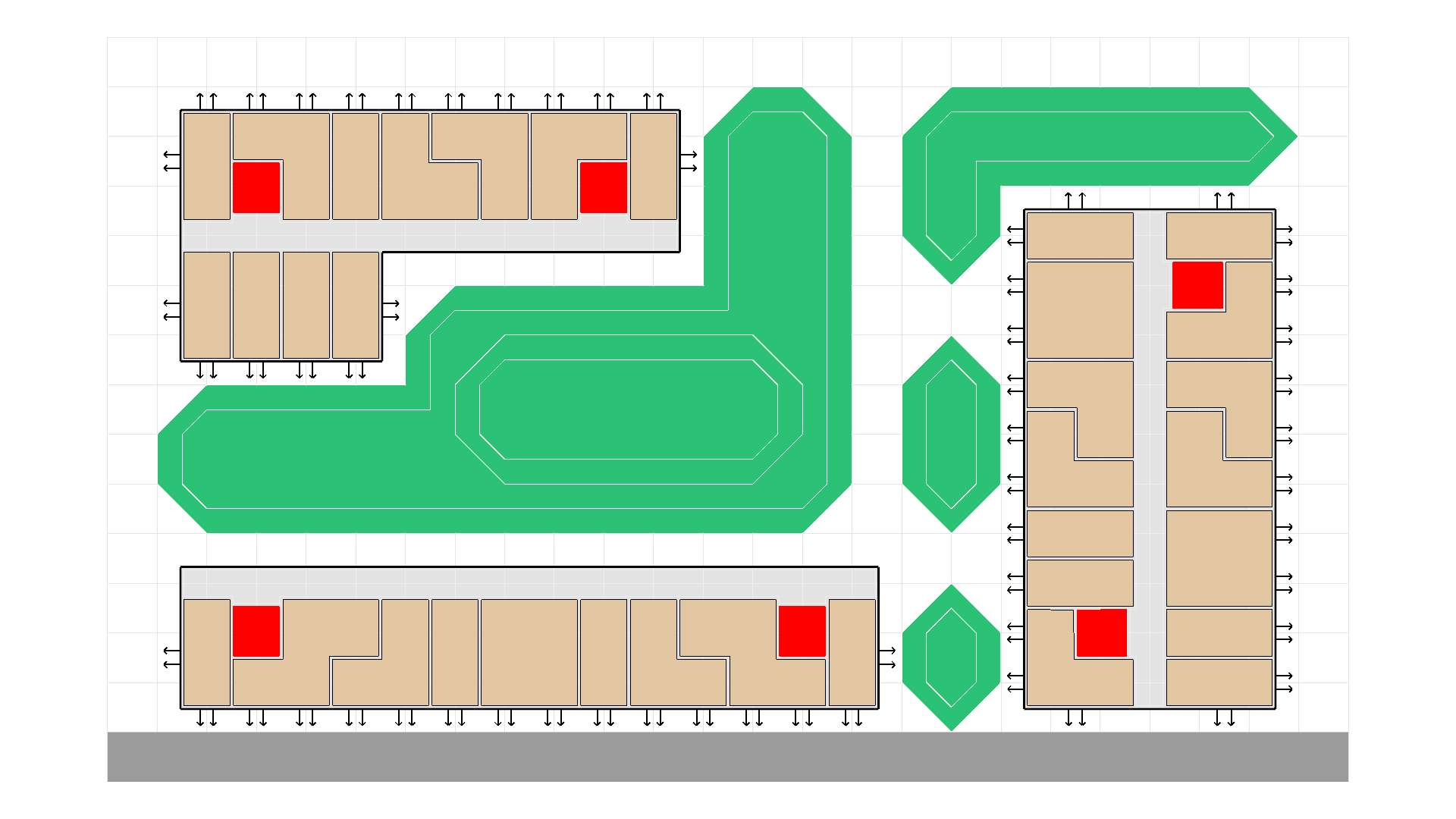} &
        \includegraphics[width=0.18\linewidth]{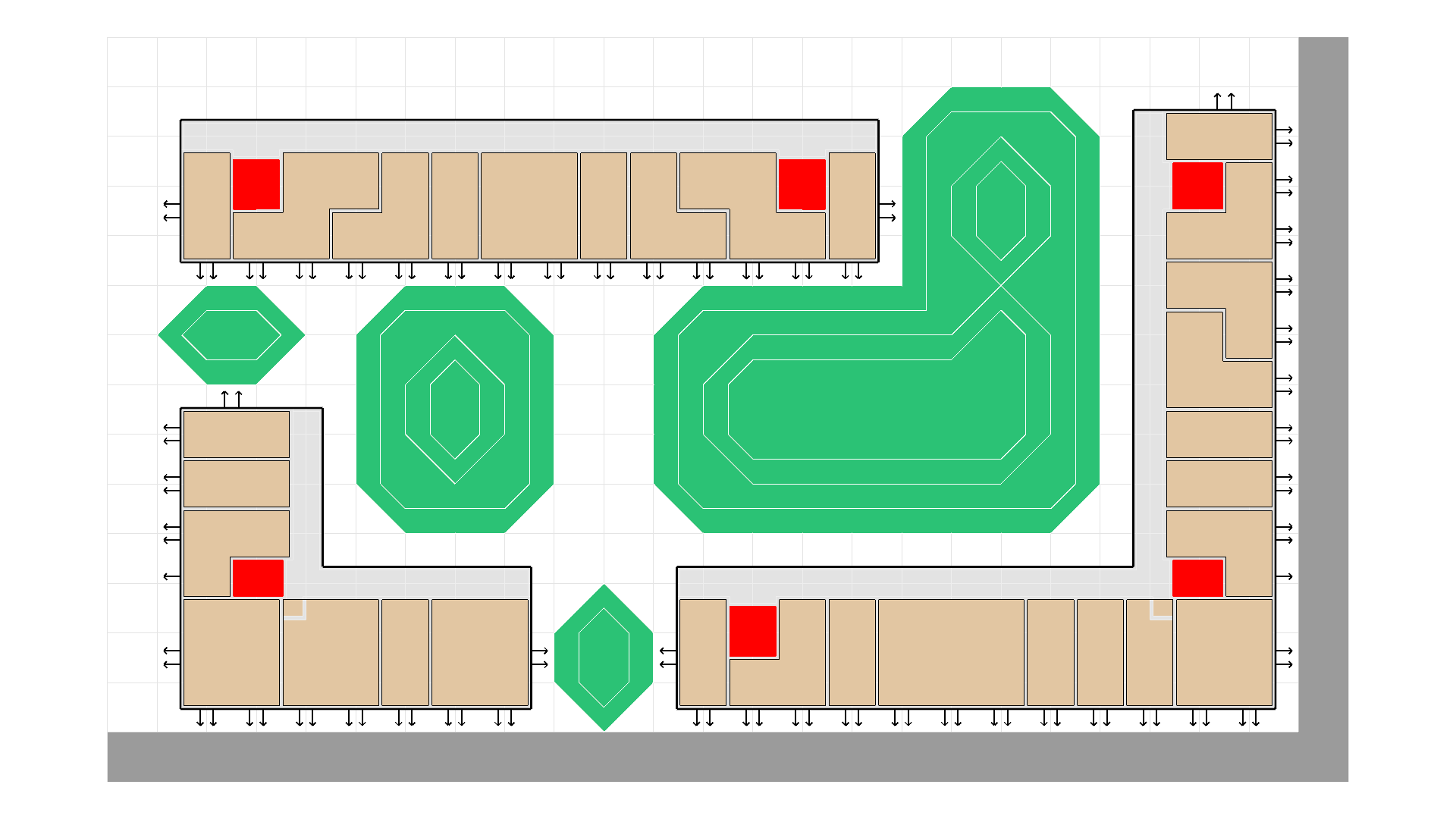} &
        \includegraphics[width=0.18\linewidth]{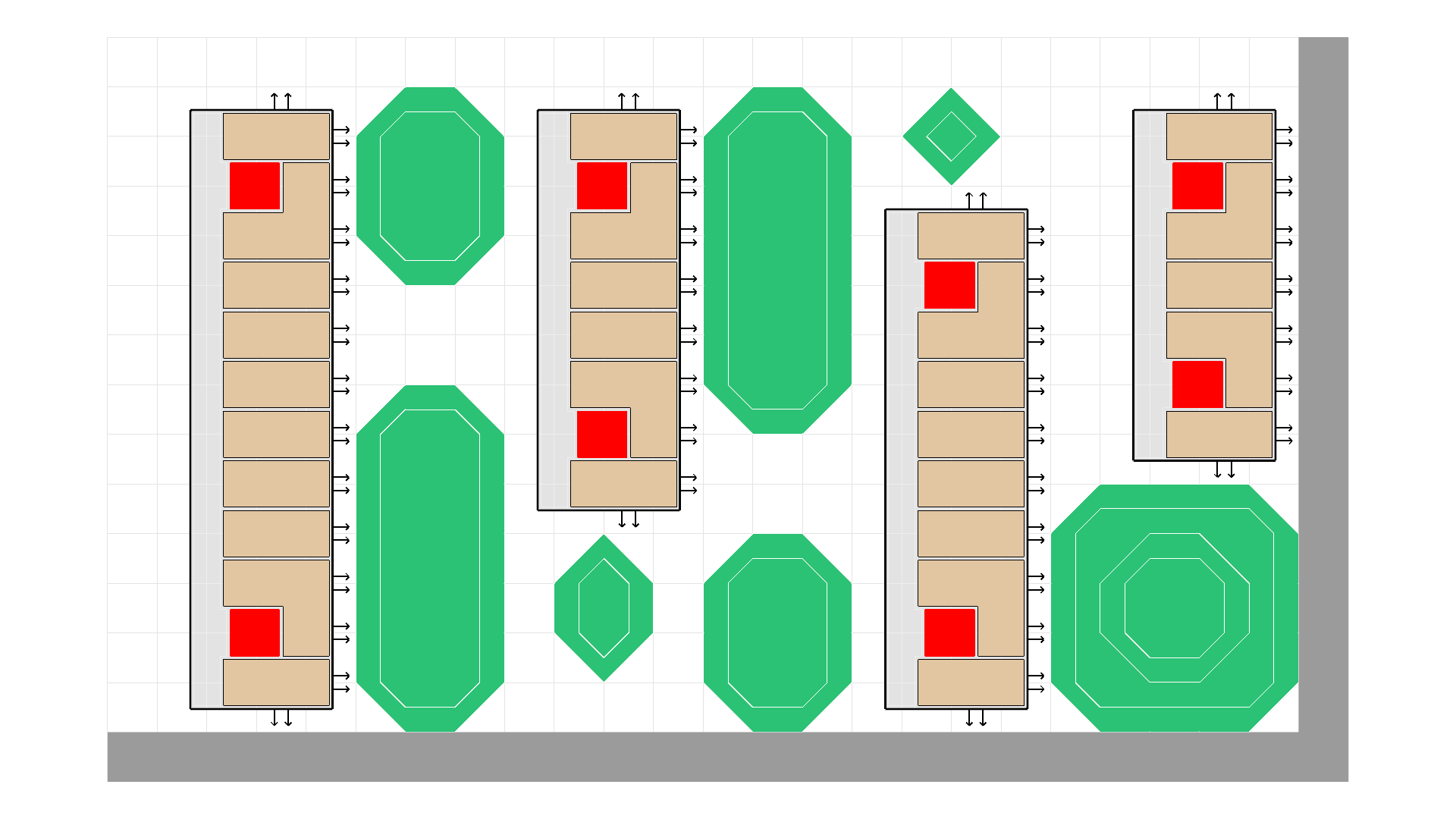} & 
        \includegraphics[width=0.18\linewidth]{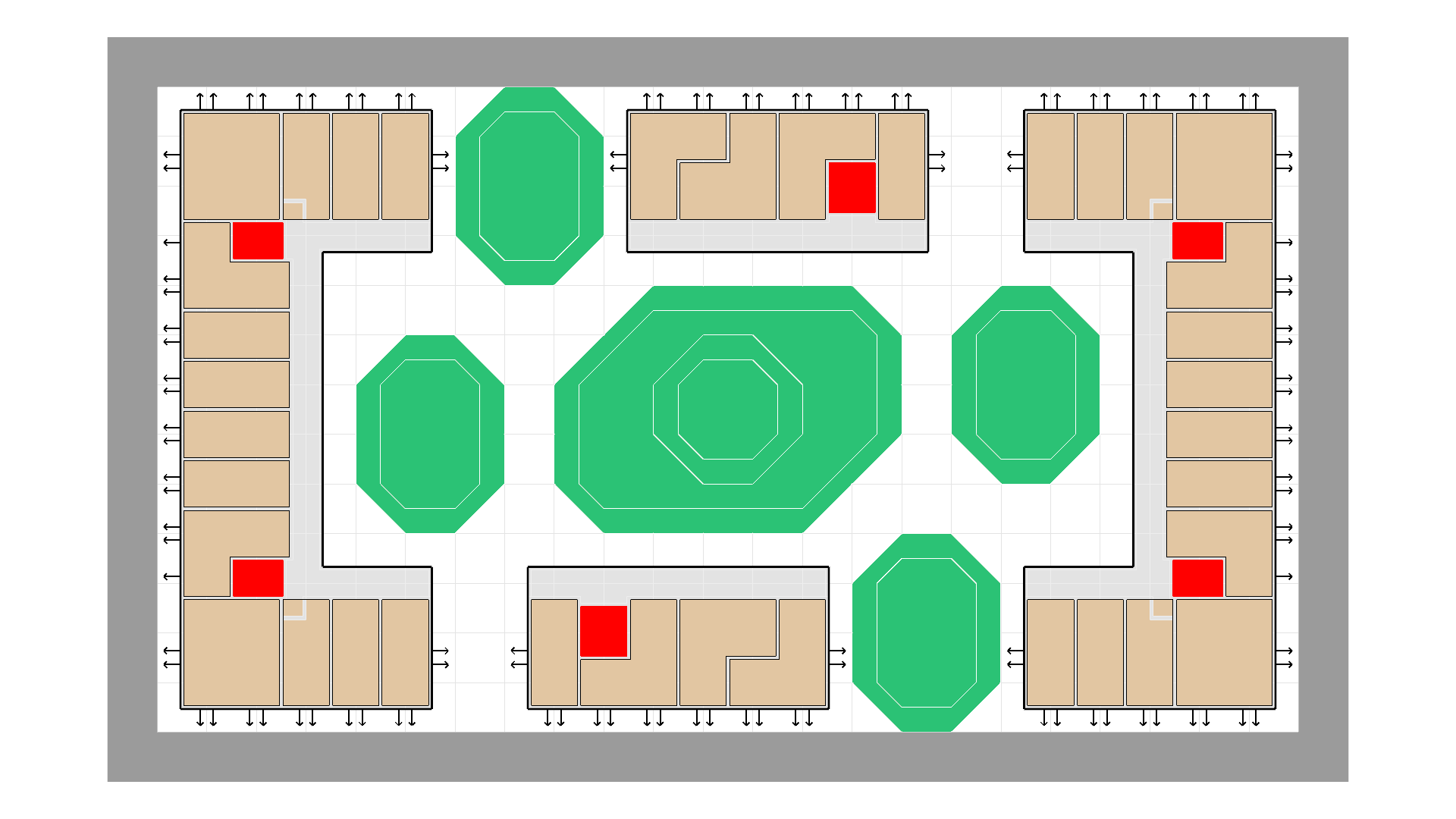} \\
    \end{tabular}
    \caption{Base designs used by Wave Function Collapse. WFC learns a set of adjacency rules from a small set of examples. In this work, which generated 50,000 different samples for two different datasets, the generator was based off the adjacency rules learned from just the above five layouts.}
    \label{table:images}
    \end{table*}

\section{Full Tile to Tile to Character Encoding}
\label{sec:tile_encoding}
    \begin{figure}[h!]
      \includegraphics[width=0.7\columnwidth]{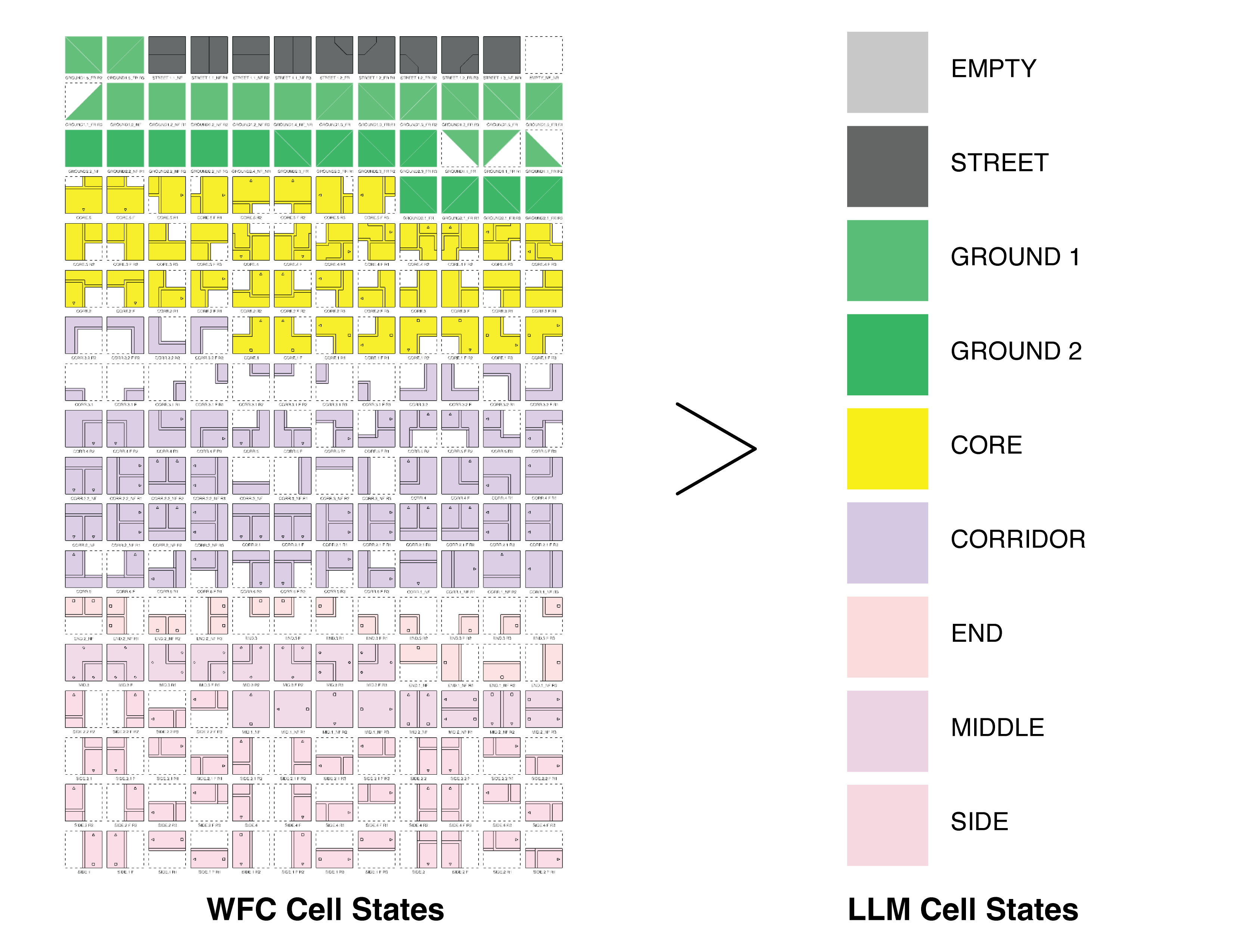}
      \caption
      { 
      Complete set of possible tile states used in WFC. 216 tile states in total, including rotations and reflections, are derived from the WFC samples. When fixed during evolution, or fed to the language mode as tokens, these states are simplified to their functional category on the right.
      }
      \label{fig:tile_encoding}
    \end{figure}

\end{document}